\documentclass[11pt]{article}
\usepackage[final]{acl}
\usepackage{times}
\usepackage[T1]{fontenc}
\usepackage[utf8]{inputenc}
\usepackage{microtype}
\usepackage{inconsolata}
\usepackage{graphicx}
\usepackage{booktabs}
\usepackage{CJKutf8}
\usepackage{amsmath}

\title{What If Chinese Were Latinized? A Counterfactual Study of Input Representation, Tokenization, and Language Modeling}

\author{
  Zijie Zheng \\
  The Hong Kong University of \\
  Science and Technology \\
  \texttt{zzhengbn@connect.ust.hk} \\\And
  Ej Zhou \\
  University of Cambridge \\
  \texttt{yz926@cam.ac.uk}
}

\begin{document}
\maketitle

\begin{abstract}

{\emph{If Chinese had been Latinized, as the Latinxua Sin Wenz movement proposed, what would it have meant for Chinese NLP?}} Homophones and tone marking complicate the answer.
We convert Chinese corpora to syllable-separated Pinyin and train superBPE tokenizers from scratch. 
The resulting Pinyin vocabularies not only exhibit expected homophone collisions, but also show approximately $10\%$ higher fertility than characters.
An alternative BPE configuration changes the relative fertility of the two representations.
We further pretrain Llama-style models on Chinese characters and Pinyin. The Pinyin models perform worse on per-character perplexity, homophone disambiguation, and overall ZhoBLiMP minimal-pair accuracy.
An exploratory phonological probe yields near-chance performance for both representations.
The broader implication is that written input representations are not merely neutral encodings. Choices for representations affect language modeling and their downstream performances. Latinization never actually happened for Chinese, but other languages did, such as Hangul and Latin script replaced character in Korean and Vietnamese; the cost we measure is a guide to where modeling effort for those languages should concentrate.

\end{abstract}

\section{Introduction}
\label{sec:intro}

In the 1920s--50s, the \textbf{Latinxua Sin Wenz} (\begin{CJK}{UTF8}{gbsn}拉丁化新文字\end{CJK}, ``Latinized New Script'') movement proposed abolishing Chinese characters in favor of a Latin alphabet.
It was championed by figures from Qu Qiubai to Lu Xun and over 600 writers and scholars \citep{defrancis1950nationalism}.\footnote{Latinxua differs from earlier Latinizations (Wade--Giles, Gwoyeu Romatzyh) as a \emph{complete} writing system rather than a pronunciation aid, with no tone marks and explicit word spacing; the Dungan language \citep{dyer1967soviet}, a Cyrillic-script Chinese variety in Central Asia, is a real fully alphabetic Chinese variety. See \citet{defrancis1984chinese,norman1988chinese,seybolt1979language,zhou1965reform,cheng1986contradictions} for broader histories.}
Although the movement ultimately failed, the question remains: \emph{how would Chinese NLP be different if Chinese had been Latinized?}

\begin{figure}[t]
    \centering
    \includegraphics[width=1\linewidth,trim=10 50 10 64,clip]{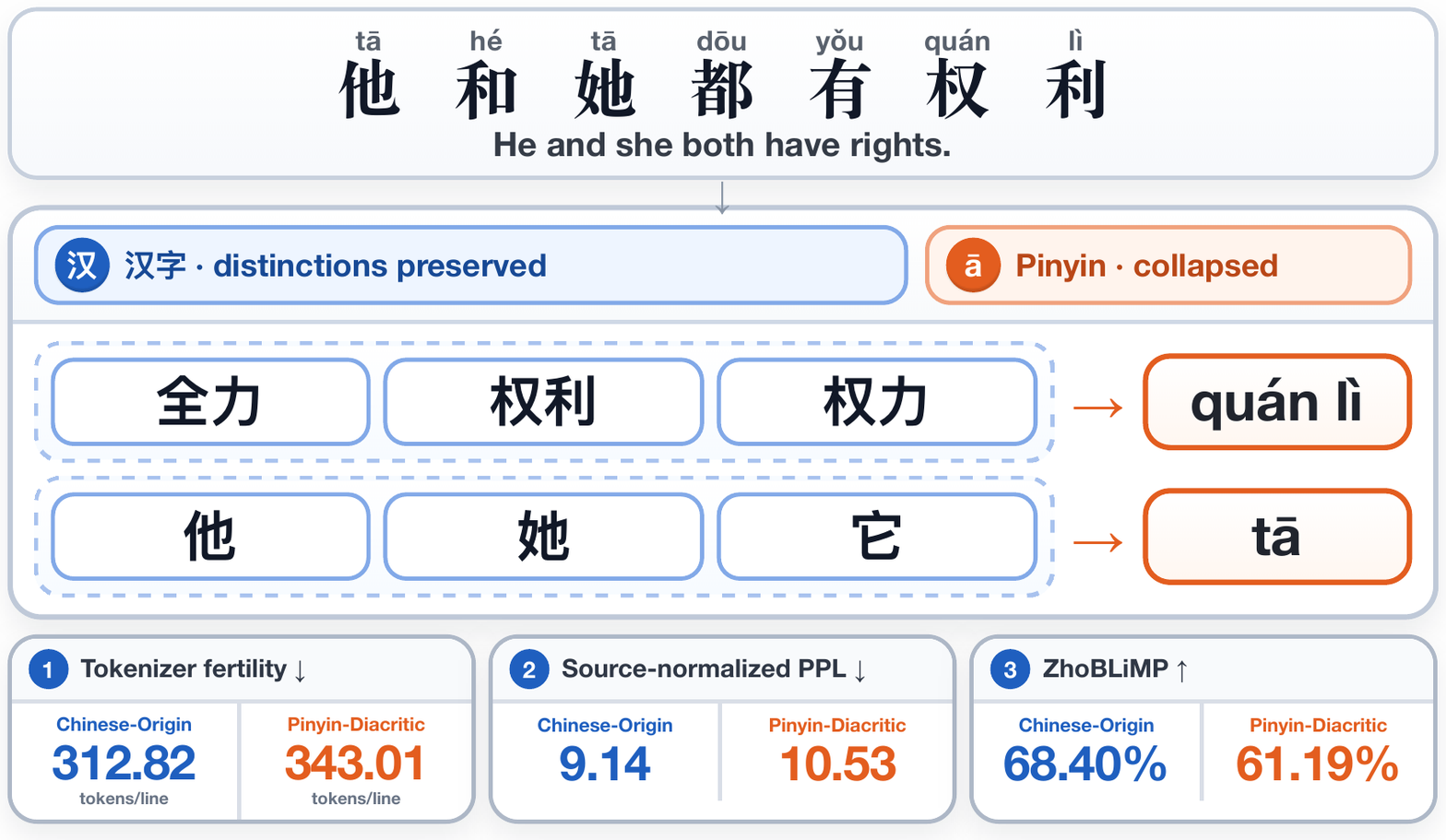}
    \caption{Overview of the counterfactual. \textbf{Top:} example homophone collisions under Latinization. \textbf{Bottom:} one tokenizer metric and two model-evaluation metrics under matched-token compute.}
    \label{fig:overview}
\end{figure}

Distinction between script and representation is important \cite{gorman-sproat-2023-myths}. We frame this as an input representation-level counterfactual: we hold the underlying Chinese content fixed and replace the character  with written Pinyin representations.
Both representations have their relative advantages.
Chinese is highly \textbf{homophonous}, and characters distinguish many homophonous morphemes orthographically,
while Pinyin represents \textbf{phonological information} more explicitly than characters, with tone-marked variants also marking lexical tones directly (\emph{shī, shí, shǐ, shì} differ by one mark).

Our experiment is illustrated in Figure~\ref{fig:overview}.
We construct our corpus based on today's standard Latin transcription of Mandarin under three orthographic conventions (toneless, number-toned, and diacritic) and train matched superBPE \citep{liu-etal-2025-superbpe} tokenizers from scratch on each.
Under both matched-token budget and matched-source control, we pretrain Llama-style language models on Chinese-Origin and Pinyin-Diacritic, and evaluate the models on held-out perplexity, a homophone-interference probe, ZhoBLiMP \citep{liu-etal-2026-systematic}, a Chinese minimal-pair acceptability benchmark, and an exploratory phonological probe. We also conduct a tokenizer-only ablation.

\paragraph{What we find.}
At the tokenizer level, Pinyin-Diacritic exhibits widespread homophone collisions and  higher fertility.
With an alternative BPE configuration, Pinyin-Diacritic instead has slightly lower fertility at 64K.
At the model level, Pinyin models perform worse on per-character perplexity, homophone disambiguation, and overall ZhoBLiMP accuracy.
The phonological probe yields near-chance results across all model conditions.
These results suggest that Chinese characters preserve lexical distinctions beyond visual information.
More broadly, input representation shapes what linguistic information is available to models, motivating representation-aware multilingual NLP and future tests of character-derived information in languages such as Korean.

\section{Related Work}
\label{sec:related}
Chinese NLP defaults to characters, but several  lines of research explore how to incorporate additional lexical, phonological, or visual information:
glyph-based models use visual features of characters \citep{meng-etal-2019-glyce}, while ChineseBERT augments character representations with both glyph and pinyin embeddings \citep{sun-etal-2021-chinesebert}.
These approaches treat pinyin as an auxiliary signal that supplements the character channel.
We instead replace Chinese characters with Pinyin, examining how the resulting differences in orthographic and phonological information relate to tokenization and language modeling. Methodologically, we follow controlled studies of separately trained
language models \citep{kallini-etal-2024-mission, zhou-etal-2026-cross}, which use synthetic transformations of English to probe what structures language models can learn.
To our knowledge, no prior NLP work uses Latinized Chinese as a counterfactual experimental framework.
\section{Tokenizer Analysis}
\label{sec:exp1}
\subsection{Setup}
\label{sec:exp1-setup}

\paragraph{Corpus and variants.}
From cleaned Chinese Wikipedia text, we construct three Pinyin orthographic variants that differ only in how tone is represented: \textbf{Pinyin-Toneless} (tones removed, e.g., \emph{zhong guo ren min}), \textbf{Pinyin-Toned} (tones as numbers, \emph{zhong1 guo2 ren2 min2}), and \textbf{Pinyin-Diacritic} (tones as diacritics, \emph{zhōng guó rén mín}); the character baseline (Chinese-Origin) is unchanged.
Spaces mark syllable boundaries rather than word boundaries.
Full preprocessing, segmentation, and splits are in Appendix~\ref{sec:appendix-pipeline}.

\paragraph{Tokenizer matrix.}
We train four vocabulary sizes per variant (8K, 16K, 32K, 64K), yielding 16 superBPE tokenizers. Tokenizer training uses a fixed 100K-line subset of the training pool; fertility evaluation uses the full held-out split.

\paragraph{SuperBPE training.}
We train superBPE \citep{liu-etal-2025-superbpe} with two-stage pre-tokenization (transition at $K{=}2048$); 
for example, superBPE can learn \emph{xī} and
\emph{wàng} separately in its first stage and
merge across the space to form \emph{xī wàng}
in its second stage.
Two-stage rationale, regexes, and robustness to alternative $K$ are reported in Appendix~\ref{app:tokenizer-training}.

\paragraph{Result.}
The vocabulary composition of the 16 tokenizers is shown in Appendix~\ref{app:additional-vocab-composition}.
\subsection{Cross-Script Token Overlap}
\label{sec:exp1-overlap}

Do the same lexical units appear as tokens in both the character-based and Latinized vocabularies?
Aligning the 64K Chinese-Origin and Pinyin-Diacritic tokenizers, we map every Chinese token to its Pinyin-Diacritic form and group by the resulting mapping multiplicity (Table~\ref{table:homophone-overlap-summary}): a 1:1 mapping is a bijective match, while $N{:}1$ mappings ($N{\geq}2$) constitute homophone collisions.

\begin{table}[t]
\centering
\small
\begin{tabular}{lrr}
\toprule
\textbf{Mapping} & \textbf{Chinese} & \textbf{Pinyin-Diacritic} \\
\midrule
Shared 1:1       & 32,025 & 32,025 \\
Shared 2:1       & 3,722  & 1,861  \\
Shared 3:1       & 1,020  & 340    \\
Shared 4:1       & 580    & 145    \\
Shared $>$ 4:1   & 3,027  & 362    \\
\bottomrule
\end{tabular}
\caption{Cross-script token overlap between the 64K tokenizers. Rows report mapping multiplicity. For example, Shared 2:1 denotes two distinct Chinese tokens mapping to one Pinyin-Diacritic token.}
\label{table:homophone-overlap-summary}
\end{table}

\paragraph{Homophone collisions.}
Table~\ref{table:homophone-overlap-summary} shows that homophone collisions are widespread.
We use \emph{collapse} to describe such \(N\):1 mappings: for example, the Pinyin-Diacritic form \emph{quán lì} collapses \begin{CJK}{UTF8}{gbsn}全力\end{CJK} (``with full effort''), \begin{CJK}{UTF8}{gbsn}权利\end{CJK} (``rights''), and \begin{CJK}{UTF8}{gbsn}权力\end{CJK} (``power'') into a single surface form.
The largest collision has multiplicity 41:1; the full distribution and collision examples are shown in Figure~\ref{figure:homophone-collision}.
Additional details and overlap results for other Pinyin tokenizers are reported in Appendix~\ref{app:additional-overlap-results}.

\paragraph{Semantic dispersion of collisions.}

Some collision groups erase grammatical or referential distinctions encoded in orthography, such as \emph{tā de} = \begin{CJK}{UTF8}{gbsn}他的/她的/它的\end{CJK} (``his/her/its'').
Other groups merge lexical items that are much farther apart semantically, such as \emph{qí shì} = \begin{CJK}{UTF8}{gbsn}骑士/歧视\end{CJK} (``knight/discrimination'').
Table~\ref{table:semantic-dispersion} illustrates that
several groups have dispersion close to a random-pair baseline, indicating substantial semantic mixing within particular homophone classes. The Chinese lexical items and glosses underlying these representative groups are listed in Appendix~\ref{sec:appendix-homophones}.
Additional results for other Pinyin tokenizers are reported in Appendix~\ref{app:additional-semantic-dispersion}.

\begin{table}[h]
\centering
\small
\begin{tabular}{@{}lrrr@{}}
\toprule
\textbf{Pinyin-Diacritic token} & \textbf{N} & \textbf{BGE} & \textbf{text2vec} \\
\midrule
\texttt{tā de}        & 4   & 0.3154 & 0.3219 \\
\texttt{àn jiàn}      & 2   & 0.6486 & 0.5867 \\
\texttt{diàn shì}     & 2   & 0.6639 & 0.7092 \\
\texttt{yù shì}       & 2   & 0.6634 & 0.7626 \\
\texttt{gōng shì}     & 6   & 0.4934 & 0.5556 \\
\texttt{xíng shì}     & 5   & 0.4682 & 0.5103 \\
\texttt{qí shì}       & 4   & 0.6436 & 0.6659 \\
\midrule
1:1 baseline & 1  & 0 & 0 \\
All         & All & 0.4028 & 0.4345 \\
Random baseline       & 10K pairs & 0.6503 & 0.6237 \\
\bottomrule
\end{tabular}
\caption{Semantic dispersion of representative \(N\):1 homophone collision groups.
Distances are average pairwise cosine distances among the collapsed Chinese tokens, computed with two models.\protect\footnotemark{}
The \emph{1:1 baseline} denotes zero dispersion for a bijective mapping.
The \emph{All} row aggregates all collision groups after retaining source tokens with at least two Chinese characters.
The \emph{random baseline} is sampled from 10K random pairs of the full Chinese vocabulary.}
\label{table:semantic-dispersion}
\end{table}

\footnotetext{BAAI/bge-small-zh-v1.5 and shibing624/text2vec-base-chinese.}

\subsection{Tokenization Fertility}
\label{sec:exp1-fertility}

We next evaluate tokenization fertility, measured as tokens per aligned corpus line, on the held-out split corresponding to each tokenizer (Table~\ref{tab:fertility}).
 Chinese-Origin has lower fertility than all three Latinized variants: despite homophone collisions, Pinyin-Diacritic requires roughly 10\% more tokens per aligned line.
A case study is
provided in Appendix~\ref{app:fertility-case-study}.
This indicates that Latinization does not automatically make Chinese more compact for tokenization, possibly because superBPE can merge continuous Chinese character sequences into stable lexical chunks, whereas Pinyin must allocate merge capacity across syllable spacing, tone-marked variants, and Latin-script boundary fragments.

\begin{table}[t]
\centering
\small
\setlength{\tabcolsep}{4pt}
\begin{tabular}{@{}lr@{}}
\toprule
\textbf{Tokenizer} & \textbf{Tokens per line} \\
\midrule
Chinese SuperBPE (64K)           & 312.82 \\
Pinyin-Toned SuperBPE (64K)      & 343.02 \\
Pinyin-Toneless SuperBPE (64K)   & 327.49 \\
Pinyin-Diacritic SuperBPE (64K)  & 343.01 \\
\midrule
Qwen-7B                          & 414.69 \\
\bottomrule
\end{tabular}

\caption{Tokens per line on the held-out zh-wiki test set corresponding to each tokenizer.
Qwen-7B is evaluated on Chinese-Origin text as a production-tokenizer reference.\protect\footnotemark}
\label{tab:fertility}
\end{table}
\footnotetext{We use the tokenizer from the Hugging Face \texttt{Qwen/Qwen-7B} base model repository.}

\paragraph{Alternative tokenizer configuration.}
Initializing BPE from bytes is a convention inherited from English and UTF-8.
A tokenizer designed for Chinese from the start could instead begin from the smallest units of the written language: characters, or syllables under Pinyin.
We therefore conduct an ablation using BPE initialized with individual Chinese characters and complete tone-marked
Pinyin syllables.
Revisiting the earlier example, this alternative BPE
starts directly with \emph{xī} and \emph{wàng}
as atomic units, allowing cross-syllable merges
from the beginning.
Standard Mandarin has about 1,300 distinct syllables when tones are counted \citep{duanmu2007phonology}, against 8,105 characters in the official Table of General Standard Chinese Characters \citep{moe2013hanzi}.
Under this configuration, Pinyin-Diacritic has lower fertility than Chinese-Origin; starting from syllables gives a more compact tokenization than starting from characters (Table~\ref{tab:syllable-bpe-results}).
Implementation details are provided
in Appendix~\ref{app:syllable-bpe-ablation}.

\begin{table}[t]
\centering
\small
\setlength{\tabcolsep}{3pt}
\begin{tabular}{@{}lrrr@{}}
\toprule
Configuration & Chinese & Pinyin & Gap \\
\midrule
superBPE 64K        & 312.82 & 343.01 & $+9.65$\%  \\
\midrule
Initialized BPE 16K & 407.21 & 353.91 & $-13.09$\% \\
Initialized BPE 32K & 323.84 & 309.48 & $-4.43$\%  \\
Initialized BPE 64K & 285.27 & 278.48 & $-2.38$\%  \\
\midrule
BPE, matched merges & 285.27 & 282.44 & $-0.99$\%  \\
\bottomrule
\end{tabular}
\caption{Fertility \(F\) in tokens per aligned held-out line for
Chinese-Origin and Pinyin-Diacritic.
Gap is \(100(F_{\mathrm{Pinyin}}/F_{\mathrm{Chinese}}-1)\%\).
At 51,122 merges, the final vocabularies contain 64,000 Chinese
entries and 57,892 Pinyin entries.}
\label{tab:syllable-bpe-results}
\end{table}

\section{Pretraining and Downstream Evaluation}
\label{sec:exp2}
In \S\ref{sec:exp1} we compared tokenization under Chinese-character and Pinyin representations.
A natural next question is whether these representational differences translate into measurable downstream consequences for language model behavior.
We therefore conduct a controlled \emph{light pretraining} study, training small language models from scratch under matched-token and matched-source settings for the two input representations.

\subsection{Experimental Setup}
\label{sec:exp2-setup}

\paragraph{Representation variants and model.} 
We restrict the comparison to \textbf{Chinese-Origin} and \textbf{Pinyin-Diacritic}.
Pinyin-Diacritic preserves tone information and is closest to the standard tone-marking convention of Pinyin.
All pretraining runs use the same 134M-parameter Llama-style decoder-only architecture (12 layers, context length 1024) and matching 32K superBPE tokenizers trained in \S\ref{sec:exp1}.
Full architecture, corpus, seed, and training-schedule details are in Appendix~\ref{app:exp2-architecture}.

\paragraph{Matching.}
We report three settings: \textbf{Chinese}, the Chinese-Origin baseline; \textbf{Pinyin-token}, the Pinyin-Diacritic model under the same matched-token budget; and \textbf{Pinyin-source}, a matched-source Pinyin-Diacritic control.
The matched-token comparison fixes the number of model-token predictions (1.781B tokens per model), matching the dominant training compute and the effective cost paid during training.
Because Pinyin has higher fertility, Pinyin-token sees fewer passes over the underlying source corpus than Chinese.
Pinyin-source instead has equal source-corpus exposure (approximately 4.0 passes, using 1.951B tokens), making it a strictly more favorable control than Pinyin-token.
All settings use three independent seeds and report mean $\pm$ standard deviation.
\subsection{Downstream Evaluation}

\label{sec:exp2-eval}
Figure~\ref{fig:exp2-loss} shows the loss curves for one seed of each setting.
We evaluate three primary aspects of model behavior motivated by the tokenizer analysis in \S\ref{sec:exp1}, followed by an exploratory phonological probe.
When a benchmark contrast collapses under Pinyin-Diacritic conversion and input becomes indistinguishable, we mark the subset after removing collapsed items with an asterisk (*).

\paragraph{Held-out perplexity.}
We evaluate
source-character-normalized
perplexity by summing the negative log-likelihood
under each model's own tokenization of the held-out
zh-wiki test split and normalizing by the number
of source Chinese characters.
For representation $r$, let $N_r$ denote the total
negative log-likelihood, $T_r$ the number of model
tokens, and $C$ the number of Unicode code points in the source text, excluding line endings. Then,
\begin{equation}
\mathrm{PPL}_{\mathrm{src}}^{(r)}
=
\exp\left(
\frac{T_r}{C}
\cdot
\frac{N_r}{T_r}
\right).
\end{equation}
Here, $T_r/C$ measures tokenization fertility,
while $N_r/T_r$ is the token-level NLL.
Source-normalized
perplexity is $9.14 \pm 0.01$ for Chinese-Origin,
compared with $10.53 \pm 0.03$ for Pinyin-token
and $10.36 \pm 0.03$ for Pinyin-source.
Thus, under matched-token compute,
Pinyin representation
increases perplexity by roughly $15.1\%$.

\paragraph{Homophone-interference lexical probe.}
We next ask whether the homophone collisions observed at the tokenizer level make lexical choice harder for Pinyin models.
We construct a 1{,}000-item probe from held-out zh-wiki text where the gold word is paired with a distractor that shares the same Pinyin after tone marks are removed, as \textbf{Pinyin Overlap}.
For example, in a sentence describing a metro station,
the probe contrasts \begin{CJK}{UTF8}{gbsn}车站\end{CJK}
(\emph{chē zhàn}, ``station'') with
\begin{CJK}{UTF8}{gbsn}车展\end{CJK}
(\emph{chē zhǎn}, ``auto show''),
which share the same toneless Pinyin.
Additionally, we include two non-homophone controls in the same sentence contexts: \textbf{Semantic Control}, matched by semantic distance to the gold word, and \textbf{Random Control}.
A structural limitation appears before scoring: 229 of the 1{,}000 same-Pinyin items collapse.
As shown in Table~\ref{tab:eval2-homophone-controls},
the performance gap between the Chinese and Pinyin
models is largest for the distinguishable
pronunciation-overlapping pairs.
Pronunciation overlap therefore adds a modest disambiguation cost beyond the general lexical-completion gap.

\begin{table}[t]
\centering
\small
\setlength{\tabcolsep}{3pt}
\resizebox{\linewidth}{!}{%
\begin{tabular}{@{}lccc@{}}
\toprule
\textbf{Probe}
& \textbf{Chinese} 
& \textbf{Pinyin-token} 
& \textbf{Pinyin-source} \\
\midrule
Pinyin overlap*
& $\mathbf{96.11}{\scriptstyle \pm 0.34}$ 
& $88.85{\scriptstyle \pm 0.72}$ 
& $88.85{\scriptstyle \pm 1.06}$ \\
Semantic control
& $\mathbf{97.33}{\scriptstyle \pm 0.35}$ 
& $90.80{\scriptstyle \pm 0.78}$ 
& $91.13{\scriptstyle \pm 0.76}$ \\
Random control
& $\mathbf{98.63}{\scriptstyle \pm 0.23}$ 
& $93.90{\scriptstyle \pm 0.46}$ 
& $93.57{\scriptstyle \pm 0.31}$ \\
\bottomrule
\end{tabular}%
}
\caption{Accuracy on the homophone-interference lexical probe.
\emph{Pinyin overlap*} excludes items collapsed by Pinyin-Diacritic conversion.
Construction and scoring details are given in Appendix~\ref{app:homophone-probe}.}
\label{tab:eval2-homophone-controls}
\end{table}

\paragraph{ZhoBLiMP.}

We evaluate ZhoBLiMP \citep{liu-etal-2026-systematic}, a Chinese BLiMP-style minimal-pair benchmark over 15 linguistic phenomena.
A model is correct if it assigns higher likelihood to the acceptable sentence than to its minimally different counterpart.
For readability, Table~\ref{tab:zhoblimp-typology} groups related phenomena.
Both Pinyin-Diacritic variants are above chance overall but remain below Chinese-Origin.
The gaps are not evenly distributed.
They are smaller for \emph{order-sensitive} phenomena such as topicalization, relativization, and passive, where Pinyin preserves sentence order.
By contrast, larger drops appear when the contrast is carried by character-distinguishable lexical or functional items, such as gendered pronouns and licensing expressions (\emph{Anaphor*} and \emph{Functional licensing}).

\begin{table}[t]
\centering
\small
\setlength{\tabcolsep}{3pt}
\resizebox{\linewidth}{!}{%
\begin{tabular}{@{}lccc@{}}
\toprule
\textbf{Phenomenon class}
& \textbf{Chinese} 
& \textbf{Pinyin-token} 
& \textbf{Pinyin-source} \\
\midrule
Functional licensing
& $\mathbf{80.0}{\scriptstyle \pm 3.1}$ 
& $50.7{\scriptstyle \pm 3.0}$ 
& $52.0{\scriptstyle \pm 5.2}$ \\

Constructional
& $\mathbf{81.6}{\scriptstyle \pm 1.2}$ 
& $71.4{\scriptstyle \pm 1.8}$ 
& $71.3{\scriptstyle \pm 1.5}$ \\

Anaphor*
& $\mathbf{57.3}{\scriptstyle \pm 1.8}$ 
& $49.9{\scriptstyle \pm 2.0}$ 
& $45.6{\scriptstyle \pm 2.3}$ \\

Argument structure
& $\mathbf{68.9}{\scriptstyle \pm 1.0}$
& $63.9{\scriptstyle \pm 0.8}$
& $64.3{\scriptstyle \pm 1.6}$ \\

Order-sensitive
& $\mathbf{77.4}{\scriptstyle \pm 2.2}$ 
& $72.5{\scriptstyle \pm 1.0}$ 
& $71.9{\scriptstyle \pm 2.3}$ \\

Unstable categories
& $45.1{\scriptstyle \pm 4.4}$ 
& $\mathbf{48.5}{\scriptstyle \pm 1.0}$ 
& $48.2{\scriptstyle \pm 1.8}$ \\

\midrule
Overall                         
& $\mathbf{68.4}{\scriptstyle \pm 0.7}$ 
& $61.2{\scriptstyle \pm 0.8}$ 
& $61.8{\scriptstyle \pm 1.3}$ \\

\bottomrule
\end{tabular}%
}

\caption{ZhoBLiMP accuracy aggregated by phenomenon class.
Class scores are macro-averages over assigned phenomena.
\emph{Anaphor*} excludes items collapsed by Pinyin-Diacritic conversion, while \emph{Unstable categories} aggregates five categories near or below chance for at least one model.
Class definitions and the full per-phenomenon breakdown are given in Appendix~\ref{app:zhoblimp-full}.}
\label{tab:zhoblimp-typology}
\end{table}

\paragraph{Phonological probe.}
To examine whether Pinyin's explicit
pronunciation information benefits phonological tasks,
we conduct an exploratory
\emph{single-character final-matching probe}.
Given a query character, models must assign higher
likelihood to a candidate sharing the same Pinyin final,
ignoring tone.
For example, given
\begin{CJK}{UTF8}{gbsn}诸\end{CJK} (\emph{zhū}),
the correct candidate is
\begin{CJK}{UTF8}{gbsn}辅\end{CJK} (\emph{fǔ})
rather than
\begin{CJK}{UTF8}{gbsn}返\end{CJK} (\emph{fǎn}),
because only the former shares the final \emph{u}.

We evaluate the models
on 600 identical two-choice items using zero-shot
and four-shot prompting.
All candidate answers are encoded as single tokens
under both the Chinese and Pinyin tokenizers.
Candidate answers are scored by conditional
log-likelihood without additional model training.
As shown in Table~\ref{tab:phonological-probe},
these near-chance results do not establish whether
input representation affects phonological capabilities.
Further details are provided in
Appendix~\ref{app:phonological-probe}.

\begin{table}[t]
\centering
\small
\setlength{\tabcolsep}{3pt}
\begin{tabular}{@{}lccc@{}}
\toprule
\textbf{Setting}
& \textbf{Chinese}
& \textbf{Pinyin-token}
& \textbf{Pinyin-source} \\
\midrule
Zero-shot
& $48.06{\scriptstyle \pm 1.40}$
& $47.61{\scriptstyle \pm 2.43}$
& $47.83{\scriptstyle \pm 1.00}$ \\
Four-shot
& $49.39{\scriptstyle \pm 0.67}$
& $49.67{\scriptstyle \pm 0.33}$
& $50.00{\scriptstyle \pm 1.45}$ \\
\bottomrule
\end{tabular}
\caption{Accuracy on the
phonological probe.
Values are mean $\pm$ standard deviation
across three training seeds.}
\label{tab:phonological-probe}
\end{table}

\section{Discussion and Conclusion}
\label{sec:discussion}
\paragraph{Representation efficiency and recoverability.}
Our tokenizer analysis reveals differences in both tokenization efficiency and the distinctions preserved by the two input representations. Chinese-Origin requires fewer tokens in our main tokenizer experiments, and the ordering reverses under the alternative configuration. Tokenization efficiency thus depends on how the written representation is tokenized.
Our lexical and grammatical evaluations reveal two distinct phenomena: \emph{representational collapse}, where Pinyin conversion makes an input contrast disappear (\begin{CJK}{UTF8}{gbsn}他\end{CJK}/\begin{CJK}{UTF8}{gbsn}她\end{CJK} both map to \emph{tā}), and overall performance gaps on distinguishable items. The near-chance results of our exploratory probe do not establish whether input representation affects phonological capabilities.

\paragraph{Beyond the Chinese case.}
Chinese was never Latinized in Pinyin as its primary written representation, but other languages provide opportunities to study alternative written representations.
Vietnamese is a tonal language written with Latin letters and tone diacritics, while Korean Hangul encodes sound in alphabetic units arranged into syllable blocks.
These cases show that representations are not neutral encodings for language models. The choice for linguistic representations shapes tokenization and recoverability.
Our experiment may motivate future controlled comparisons of written input representations, such as testing Hanja-derived information in Korean.
For multilingual LLMs, where a few languages already receive most of the
impact from language technology \citep{occhini2026hierarchy}, cross-lingual
fairness should also account for how input representations make linguistic
information more or less accessible to models.

\clearpage
\newpage

\section*{Limitations}
\label{sec:limitations}

Automated Pinyin conversion may introduce segmentation and pronunciation errors.
Our findings depend on the chosen input representations and tokenizer configurations; while our ablation shows that tokenization efficiency varies across configurations, their downstream effects remain untested.
Near-chance phonological-probe results may reflect limited model capabilities or evaluation design.
The 134M models and Wikipedia-only corpus limit generalizability to larger models and other domains.

\section*{Ethics Statement}

This study uses publicly available Chinese Wikipedia text under the CC BY-SA 4.0 license and existing linguistic evaluation resources.
No human participants were recruited, and we conduct no individual-level analysis.
Our counterfactual examines written input representations for language modeling rather than advocating orthographic reform.
We do not foresee specific risks of misuse beyond those associated
with any small-scale Chinese language model.
We used AI-assisted tools for language polishing, manuscript revisions, and figure editing.
The authors reviewed and verified the scientific content and take full responsibility for the final paper.

\bibliography{custom}

\appendix
\renewcommand{\topfraction}{0.90}
\renewcommand{\dbltopfraction}{0.90}
\renewcommand{\textfraction}{0.08}
\renewcommand{\floatpagefraction}{0.78}
\renewcommand{\dblfloatpagefraction}{0.78}
\section{Tokenizer Analysis}
\label{app:tokenizer-analysis}

\subsection{Corpus, tokenizers, and vocabularies}
\label{sec:appendix-pipeline}

\paragraph{Corpus and script conversion.}
We extract Chinese Wikipedia articles with WikiExtractor, convert traditional to simplified Chinese, remove citation markers, wiki links, and selected markup, and truncate reference and external-link sections. We normalize whitespace and map common Chinese punctuation to Western equivalents. Numbers, URLs, and existing Latin and other non-Chinese text are preserved.

We use \texttt{jieba} \citep{jieba} for word segmentation to give \texttt{pypinyin} \citep{pypinyin} context for polyphonic characters. Chinese-Origin retains its spacing. In the three Pinyin variants, each character with an available reading yields one syllable, and spaces separate adjacent syllables (Table~\ref{tab:conversion-example}). For Pinyin-Diacritic, we use \texttt{pypinyin.lazy\_pinyin} with \texttt{style=Style.TONE} and \texttt{neutral\_tone\_\allowbreak with\_five=False}.

\begin{table}[!htbp]
\centering
\small
\begin{tabular}{@{}ll@{}}
\toprule
\textbf{Variant} & \textbf{Output} \\
\midrule
Original   & \begin{CJK}{UTF8}{gbsn}中华人民共和国\end{CJK} \\
Toned      & zhong1 hua2 ren2 min2 gong4 he2 guo2 \\
Toneless   & zhong hua ren min gong he guo \\
Diacritic   & zhōng huá rén mín gòng hé guó  \\
\bottomrule
\end{tabular}
\caption{Example conversion of \begin{CJK}{UTF8}{gbsn}中华人民共和国\end{CJK} (``People's Republic of China'') under different orthographic conventions.}
\label{tab:conversion-example}
\end{table}

\paragraph{Tokenizer data split.}
We shuffle the 1,337,505 aligned line indices with seed 42, reserving
1,203,754 lines for the training pool and 133,751 for evaluation. We
sample 100,000 training lines with a fixed seed and reuse their indices
across all representations and vocabulary sizes. Fertility is evaluated
on the full held-out split.

\paragraph{Two-stage superBPE training.}
\leavevmode\phantomsection\label{app:tokenizer-training}%
We train superBPE tokenizers \citep{liu-etal-2025-superbpe}
in two stages. Stage~1 uses \verb!\S+|\s+! to separate whitespace
from non-whitespace spans. On Pinyin text, this keeps merges within
syllables. Stage~2 inherits the first \(K\) merges and uses
\verb![^\p{L}\p{N}\s]+|[\r\n]+! to isolate punctuation and line
breaks while permitting merges across syllable spaces. We set
\(K=2048\) for all four representations and vocabulary sizes
(8K, 16K, 32K, and 64K).

\paragraph{Transition-point sensitivity.}
\leavevmode\phantomsection\label{app:tpoint-robustness}%
For the two 64K tokenizers in Table~\ref{tab:tpoint-baseline-values}, we vary the number of inherited merges, \(K\), over \(\{1000,3200,6400\}\).

\begin{table}[!htbp]
\centering
\small
\setlength{\tabcolsep}{1.8pt}
\begin{tabular}{@{}lrr@{}}
\toprule
Metric & Chinese & Diacritic \\
\midrule
\multicolumn{3}{@{}l}{\emph{Baseline at \(K=2048\)}} \\
Vocabulary (unique) & 61,143 & 42,653 \\
Vocabulary (raw) & 63,499 & 63,754 \\
Shared tokens & 40,374 & 34,733 \\
Independent tokens & 20,768 & 7,919 \\
Tokens/original char. & 0.5448 & 0.5974 \\
\midrule
\multicolumn{3}{@{}l}{\emph{Largest deviation (alternative \(K\))}} \\
Vocabulary (unique) & $+0.05\%$ (1000) & $+1.46\%$ (1000) \\
Vocabulary (raw) & $+0.00\%$ (1000) & $+0.01\%$ (6400) \\
Shared tokens & $+1.69\%$ (1000) & $+1.94\%$ (1000) \\
Independent tokens & $-3.14\%$ (1000) & $+6.84\%$ (6400) \\
Tokens/original char. & $-0.02\%$ (6400) & $+0.25\%$ (6400) \\
\bottomrule
\end{tabular}
\caption{Transition-point sensitivity at 64K. The lower panel gives the signed change at the \(K\) with the largest absolute percentage change for each metric and representation. Unique counts ignore whitespace, and raw counts include each vocabulary entry.}
\label{tab:tpoint-baseline-values}
\label{tab:tpoint-max-deviation}
\end{table}

\paragraph{Vocabulary composition.}
\leavevmode\phantomsection\label{app:additional-vocab-composition}%
Tables~\ref{tab:vocab-composition} and~\ref{tab:appendix-vocab-composition-bc}
give the vocabulary composition of all 16 superBPE tokenizers.
Each cell gives the number of distinct token contents after whitespace
removal, followed by the raw entry count in parentheses. Token length
is measured in Chinese characters for Chinese-Origin and in syllables
for Pinyin.

\begin{table*}[!tp]
\centering
\scriptsize
\setlength{\tabcolsep}{0.8pt}
\begin{tabular*}{\textwidth}{@{\extracolsep{\fill}}lrrrrrrrr@{}}
\toprule
& \multicolumn{4}{c}{\textbf{Chinese-Origin}}
& \multicolumn{4}{c}{\textbf{Pinyin-Diacritic}} \\
\cmidrule(lr){2-5}\cmidrule(l){6-9}
Category & 8K & 16K & 32K & 64K & 8K & 16K & 32K & 64K \\
\midrule
Single Latin letters
& 52 (94) & 52 (104) & 52 (117) & 52 (132)
& 52 (163) & 52 (232) & 52 (301) & 52 (342) \\
Single CJK characters
& 2378 (2418) & 3061 (3181) & 3769 (4021) & 4515 (4937)
& 0 (0) & 0 (0) & 0 (0) & 0 (0) \\
Digits
& 155 (185) & 247 (295) & 312 (394) & 558 (696)
& 294 (420) & 294 (587) & 314 (826) & 397 (1215) \\
Punctuation
& 131 (131) & 201 (202) & 308 (309) & 471 (472)
& 58 (59) & 62 (65) & 70 (73) & 94 (97) \\
Pure Latin combinations
& 274 (289) & 713 (785) & 1541 (1753) & 3194 (3780)
& 1644 (3256) & 1692 (4078) & 2174 (5480) & 3341 (8003) \\
\midrule
Length 2
& 3443 (3483) & 7589 (7722) & 15000 (15320) & 27294 (28053)
& 2546 (3327) & 6044 (8418) & 11851 (17460) & 21468 (33200) \\
Length 3
& 556 (556) & 1946 (1949) & 5345 (5367) & 13200 (13293)
& 259 (296) & 1226 (1404) & 3637 (4361) & 9390 (11658) \\
Length 4
& 95 (97) & 442 (446) & 1684 (1698) & 5696 (5739)
& 50 (54) & 256 (287) & 974 (1107) & 3393 (3991) \\
Length 5
& 23 (23) & 100 (100) & 389 (389) & 1340 (1343)
& 23 (23) & 61 (64) & 247 (262) & 867 (963) \\
Length 6+
& 40 (42) & 121 (125) & 386 (395) & 1335 (1348)
& 32 (34) & 102 (106) & 302 (319) & 1026 (1098) \\
\midrule
Mixed Latin and CJK
& 5 (5) & 12 (12) & 49 (49) & 157 (161)
& 0 (0) & 0 (0) & 0 (0) & 0 (0) \\
Other / unknown
& 242 (265) & 593 (652) & 1595 (1734) & 3331 (3545)
& 206 (241) & 469 (623) & 1346 (1632) & 2627 (3187) \\
\midrule
\textbf{Total}
& 7394 (7588) & 15077 (15573) & 30430 (31546) & 61143 (63499)
& 5164 (7873) & 10258 (15864) & 20967 (31821) & 42653 (63754) \\
\bottomrule
\end{tabular*}
\caption{Vocabulary composition of the Chinese-Origin and
Pinyin-Diacritic superBPE tokenizers. Single Latin letters comprise
A--Z and a--z.}
\label{tab:vocab-composition}
\end{table*}

\begin{table*}[!tp]
\centering
\scriptsize
\setlength{\tabcolsep}{0.8pt}
\begin{tabular*}{\textwidth}{@{\extracolsep{\fill}}lrrrrrrrr@{}}
\toprule
& \multicolumn{4}{c}{\textbf{Pinyin-Toneless}}
& \multicolumn{4}{c}{\textbf{Pinyin-Toned}} \\
\cmidrule(lr){2-5}\cmidrule(l){6-9}
Category & 8K & 16K & 32K & 64K & 8K & 16K & 32K & 64K \\
\midrule
Single Latin letters
& 52 (168) & 52 (237) & 52 (309) & 52 (343)
& 52 (161) & 52 (234) & 52 (299) & 52 (342) \\
Single CJK characters
& 0 (0) & 0 (0) & 0 (0) & 0 (0)
& 0 (0) & 0 (0) & 0 (0) & 0 (0) \\
Digits
& 423 (556) & 423 (735) & 425 (985) & 486 (1392)
& 296 (422) & 296 (587) & 315 (825) & 398 (1217) \\
Punctuation
& 67 (68) & 67 (70) & 74 (77) & 95 (98)
& 58 (59) & 62 (65) & 70 (73) & 94 (97) \\
Pure Latin combinations
& 1447 (2410) & 1453 (2947) & 1626 (3817) & 2757 (5965)
& 1652 (3268) & 1699 (4085) & 2179 (5488) & 3357 (8005) \\
\midrule
Length 2
& 3325 (3901) & 7099 (8995) & 12817 (17313) & 20900 (30400)
& 2545 (3326) & 6043 (8417) & 11851 (17462) & 21466 (33203) \\
Length 3
& 364 (388) & 1537 (1680) & 4880 (5490) & 12888 (14811)
& 258 (295) & 1224 (1403) & 3639 (4362) & 9392 (11660) \\
Length 4
& 58 (63) & 324 (353) & 1263 (1392) & 4312 (4807)
& 50 (54) & 256 (287) & 973 (1106) & 3393 (3990) \\
Length 5
& 29 (30) & 77 (79) & 333 (344) & 1068 (1145)
& 23 (23) & 61 (64) & 248 (263) & 867 (963) \\
Length 6+
& 36 (39) & 108 (113) & 387 (400) & 1251 (1309)
& 32 (34) & 102 (106) & 303 (320) & 1026 (1098) \\
\midrule
Mixed Latin and CJK
& 0 (0) & 0 (0) & 0 (0) & 0 (0)
& 0 (0) & 0 (0) & 0 (0) & 0 (0) \\
Other / unknown
& 204 (245) & 479 (647) & 1359 (1682) & 2844 (3471)
& 198 (233) & 461 (615) & 1338 (1624) & 2619 (3180) \\
\midrule
\textbf{Total}
& 6001 (7868) & 11608 (15856) & 23197 (31809) & 46602 (63741)
& 5164 (7875) & 10256 (15863) & 20968 (31822) & 42663 (63755) \\
\bottomrule
\end{tabular*}
\caption{Vocabulary composition of the Pinyin-Toneless and
Pinyin-Toned superBPE tokenizers.}
\label{tab:appendix-vocab-composition-bc}
\end{table*}

\subsection{Cross-script token overlap}
\label{app:additional-overlap-results}

\paragraph{Mapping multiplicity and collision examples.}
\leavevmode\phantomsection\label{sec:appendix-homophones}%
Following \S\ref{sec:exp1-overlap}, we convert each token in the
64K Chinese-Origin vocabulary to its Pinyin form and look it up in
the corresponding 64K Pinyin vocabulary. An \(N\):1 mapping groups
\(N\) distinct Chinese tokens under one shared Pinyin form.
We exclude whitespace-only tokens from the overlap counts.
The vocabulary composition tables count them as one distinct
\emph{Other / unknown} item after whitespace removal.

\begin{figure*}[!tp]
\centering
\includegraphics[
    width=\textwidth,
    height=0.45\textheight,
    keepaspectratio
]{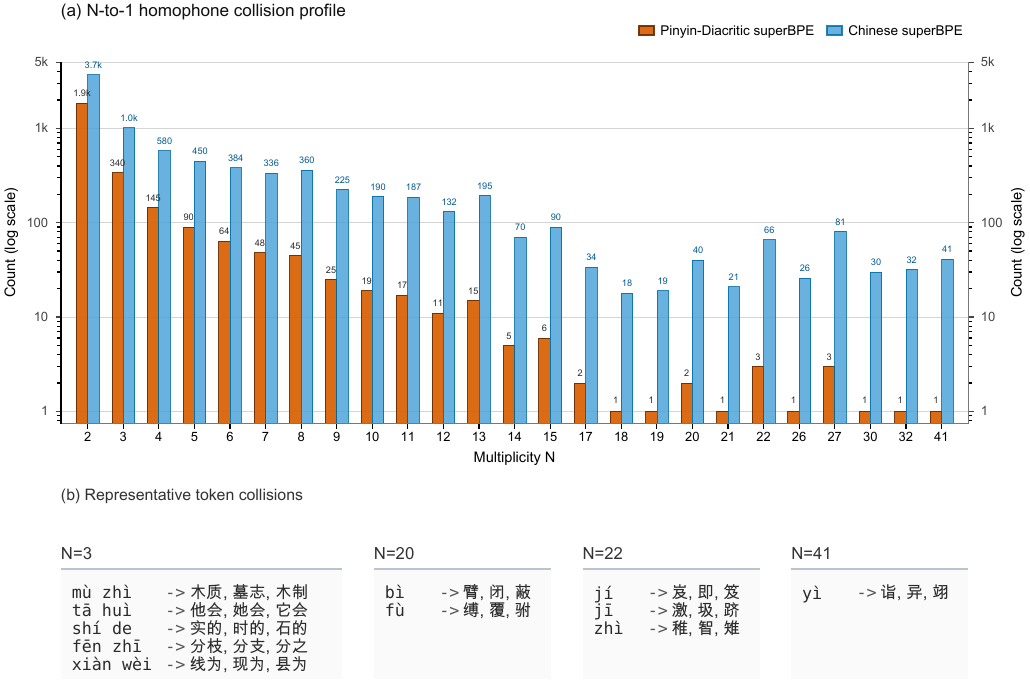}
\caption{Homophone collisions in the 64K Chinese-Origin and
Pinyin-Diacritic vocabularies. For each multiplicity \(N\), orange
bars count Pinyin forms and blue bars count the corresponding
Chinese source tokens. The lower panel gives examples.}
\label{figure:homophone-collision}
\end{figure*}

Table~\ref{table:appendix-overlap-ab} reports the Toneless and Toned counts. The maximum multiplicity is 77:1 for Toneless and 41:1 for both Toned and Diacritic. Figure~\ref{figure:homophone-collision} gives the Diacritic distribution, and Table~\ref{tab:homophone-examples} lists selected groups with English glosses.

\begin{table}[!htbp]
\centering
\small
\setlength{\tabcolsep}{3pt}
\begin{tabular}{@{}lrrrr@{}}
\toprule
& \multicolumn{2}{c}{Toneless}
& \multicolumn{2}{c}{Toned} \\
\cmidrule(lr){2-3}\cmidrule(l){4-5}
Mapping & Chinese & Pinyin & Chinese & Pinyin \\
\midrule
1:1       & 28,045 & 28,045 & 32,006 & 32,006 \\
2:1       &  6,572 &  3,286 &  3,720 &  1,860 \\
3:1       &  2,937 &    979 &  1,017 &    339 \\
4:1       &  1,484 &    371 &    584 &    146 \\
\(>4\):1  &  6,369 &    596 &  3,025 &    362 \\
\bottomrule
\end{tabular}
\caption{Cross-script overlap at 64K. Each column pair counts
the Chinese source tokens and their distinct shared Pinyin forms.}
\label{table:appendix-overlap-ab}
\label{table:appendix-overlap-ac}
\end{table}

\begin{table*}[!tp]
\centering
\small
\setlength{\tabcolsep}{4pt}
\begin{tabular}{@{}p{1.6cm}p{4.6cm}p{7.4cm}@{}}
\toprule
Pinyin & Chinese tokens & English glosses \\
\midrule
\emph{quán lì}
& \begin{CJK}{UTF8}{gbsn}全力、权利、权力\end{CJK}
& with full effort, rights, power \\

\emph{tā de}
& \begin{CJK}{UTF8}{gbsn}他的、她的、它的、\end{CJK}%
  \begin{CJK}{UTF8}{bsmi}牠的\end{CJK}
& his, her, its, its (for animals) \\

\emph{àn jiàn}
& \begin{CJK}{UTF8}{gbsn}案件、按键\end{CJK}
& legal case, button \\

\emph{diàn shì}
& \begin{CJK}{UTF8}{gbsn}电视、殿试\end{CJK}
& television, palace examination \\

\emph{yù shì}
& \begin{CJK}{UTF8}{gbsn}预示、浴室\end{CJK}
& to foreshadow, bathroom \\

\emph{qí shì}
& \begin{CJK}{UTF8}{gbsn}其事、骑士、棋士、歧视\end{CJK}
& that matter, knight, chess player, discrimination \\

\emph{gōng shì}
& \begin{CJK}{UTF8}{gbsn}公事、公式、公示、公视、\allowbreak{}工事、攻势\end{CJK}
& official business, formula, public notice, Public TV Service,
  engineering works, offensive \\

\emph{xíng shì}
& \begin{CJK}{UTF8}{gbsn}刑事、型式、形势、形式、\allowbreak{}行事\end{CJK}
& criminal, type, situation, form, conduct affairs \\
\bottomrule
\end{tabular}
\caption{Selected collision groups in the 64K Pinyin-Diacritic
vocabulary. Glosses follow the order of the Chinese tokens.}
\label{tab:homophone-examples}
\end{table*}

\paragraph{Semantic dispersion.}
\leavevmode\phantomsection\label{app:additional-semantic-dispersion}%
For each collision group, we compute the mean pairwise cosine
distance between the embeddings of its Chinese tokens, separately
for BGE and text2vec. The aggregate row includes source tokens
with at least two CJK characters. The random baseline uses 10,000
token pairs sampled from the full Chinese vocabulary.

Table~\ref{table:appendix-semantic-dispersion-toneless} gives the
Toneless and Toned results.

\begin{table*}[!tp]
\centering
\small
\setlength{\tabcolsep}{5pt}
\begin{tabular}{@{}lrrrlrrr@{}}
\toprule
\multicolumn{4}{c}{\textbf{Pinyin-Toneless}}
& \multicolumn{4}{c}{\textbf{Pinyin-Toned}} \\
\cmidrule(lr){1-4}\cmidrule(l){5-8}
Token & \(N\) & BGE & text2vec
& Token & \(N\) & BGE & text2vec \\
\midrule
\texttt{ta de}    &  5 & 0.3930 & 0.4085
& \texttt{ta1 de}      & 4 & 0.3154 & 0.3219 \\
\texttt{an jian}  &  3 & 0.5895 & 0.6556
& \texttt{an4 jian4}   & 2 & 0.6486 & 0.5867 \\
\texttt{dian shi} &  3 & 0.6108 & 0.6449
& \texttt{dian4 shi4}  & 2 & 0.6639 & 0.7092 \\
\texttt{yu shi}   &  8 & 0.6051 & 0.6227
& \texttt{yu4 shi4}    & 2 & 0.6634 & 0.7626 \\
\texttt{gong shi} & 11 & 0.5304 & 0.5649
& \texttt{gong1 shi4}  & 6 & 0.4934 & 0.5556 \\
\texttt{xing shi} &  8 & 0.5039 & 0.5786
& \texttt{xing2 shi4}  & 5 & 0.4682 & 0.5103 \\
\texttt{qi shi}   & 13 & 0.5927 & 0.5931
& \texttt{qi2 shi4}    & 4 & 0.6436 & 0.6659 \\
\midrule
1:1 baseline & 1 & 0 & 0
& 1:1 baseline & 1 & 0 & 0 \\
All groups & & 0.4831 & 0.4991
& All groups & & 0.4026 & 0.4343 \\
Random pairs & & 0.6505 & 0.6224
& Random pairs & & 0.6500 & 0.6231 \\
\bottomrule
\end{tabular}
\caption{Semantic dispersion for the 64K Pinyin-Toneless and
Pinyin-Toned vocabularies. \(N\) is the number of Chinese tokens
in the group for the listed Pinyin form.}
\label{table:appendix-semantic-dispersion-toneless}
\label{table:appendix-semantic-dispersion-toned}
\end{table*}

\subsection{Tokenization fertility}
\label{app:fertility-case-study}

\paragraph{Segmentation example.}
The two 64K superBPE tokenizers segment the same corpus excerpt
as follows. Vertical bars mark token boundaries.

\noindent\textbf{Chinese-Origin (11 tokens).}
\begin{CJK}{UTF8}{gbsn}
第一次世界大战 | 后由 | 伍德 | 罗 | 威尔逊 | 为 | 国际联盟 | 成立的 | 合理 | 化 | 开始
\end{CJK}

\par\noindent\textbf{Pinyin-Diacritic (15 tokens).}
dì | yí cì shì jiè dà | zhàn hòu | yóu |
wǔ dé | luó | wēi ěr | xùn wèi | guó jì |
lián méng | chéng lì de | hé | lǐ | huà | kāi shǐ

\paragraph{Character- and syllable-initialized BPE.}
\leavevmode\phantomsection\label{app:syllable-bpe-ablation}%
We train BPE tokenizers from scratch using individual Chinese
characters or complete tone-marked Pinyin syllables as initial units.
For Pinyin, a syllable with a leading space and the same syllable
without one are distinct initial symbols. Including byte-fallback
symbols, the initial vocabularies contain 12,878 Chinese entries
and 6,770 Pinyin entries. We apply frequency-based merges without
whitespace or punctuation pre-tokenization.

We reuse the 100,000 aligned tokenizer-training lines and measure
fertility on all 133,751 held-out lines. Table~\ref{tab:syllable-bpe-results}
compares the tokenizers at final vocabulary sizes of 16K, 32K, and
64K. The Chinese initial vocabulary already exceeds 8K. Because
equal final vocabulary sizes allow different numbers of learned
merges, we also compare the tokenizers after 51,122 merges each.
Only the tokenizers are retrained in this ablation.

\section{Pretraining and Downstream \mbox{Evaluation}}
\label{app:pretraining-evaluation}

\subsection{Pretraining setup}
\label{app:exp2-architecture}

\paragraph{Architecture.}
All runs use a Llama-style causal language model with 12 layers,
12 attention heads, hidden size 768, intermediate size 2048, and
a context length of 1024 tokens. The model uses SiLU activations,
RMSNorm, and rotary positional embeddings. Each tokenizer has
32,001 entries, including a shared EOS/PAD token (ID 32,000).
With untied input and output embeddings, each model has
134,107,392 trainable parameters. We train on an NVIDIA A100 GPU.

\paragraph{Corpus and batching.}
For language-model pretraining, we repartition the aligned corpus
from Appendix~\ref{sec:appendix-pipeline} into 1,310,755 training
lines and 13,375 lines each for validation and test. Packing the
training split into 1024-token blocks yields 434,846 Chinese-Origin
blocks and 476,236 Pinyin-Diacritic blocks. With 16 blocks per batch
and four gradient accumulation steps, each optimizer update uses
65,536 tokens. After dropping incomplete batches, one effective pass
contains 6,794 updates (445,251,584 tokens) for Chinese-Origin and
7,441 updates (487,653,376 tokens) for Pinyin-Diacritic.

\begin{figure}[t]
\centering
\includegraphics[width=\linewidth]{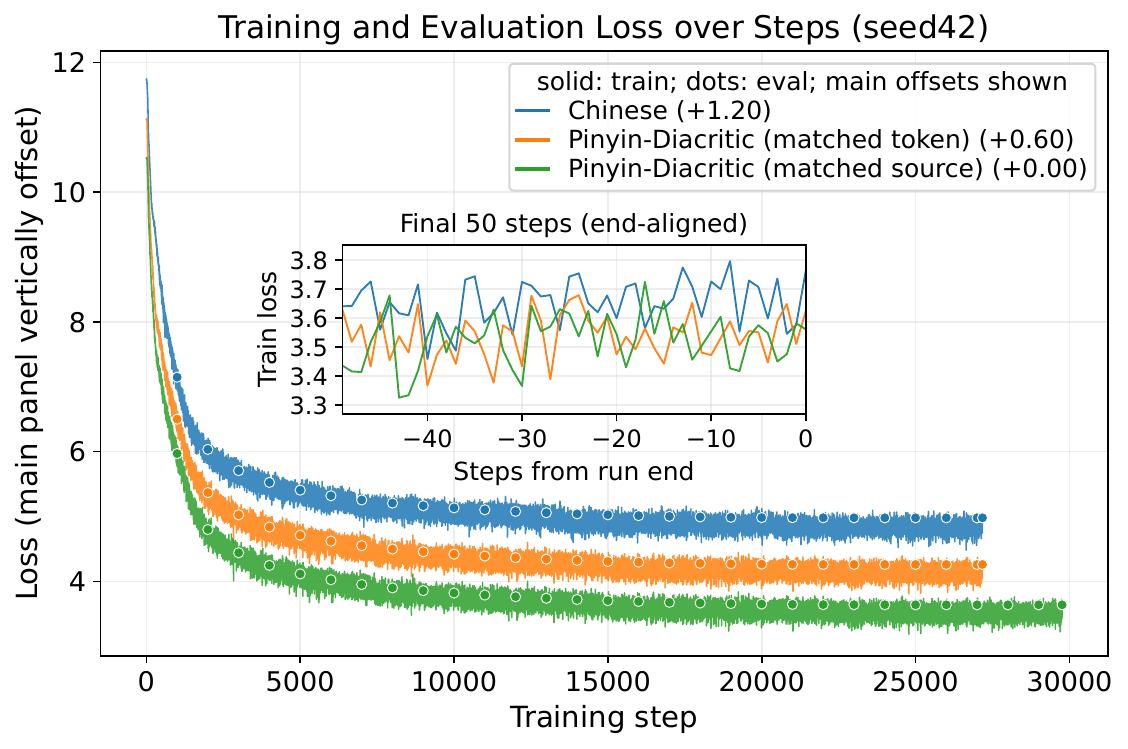}
\caption{Training and validation loss per token under each model's
own tokenizer (seed 42). The main panel uses vertical offsets for
readability. The inset aligns the final 50 updates of each run.}
\label{fig:exp2-loss}
\end{figure}

\paragraph{Training conditions.}
Table~\ref{tab:pretraining-budgets} gives the budget per run.
Pinyin-token matches the Chinese baseline in training tokens,
and Pinyin-source matches it in effective passes over the aligned
source lines. We train each condition with seeds 42, 43, and 44
and report the mean and sample standard deviation across seeds.
Figure~\ref{fig:exp2-loss} shows the token-level loss curves for
seed~42.

\begin{table}[!htbp]
\centering
\small
\setlength{\tabcolsep}{3pt}
\begin{tabular}{@{}lrrr@{}}
\toprule
Condition & Updates & Training tokens & Passes \\
\midrule
Chinese      & 27,176 & 1,781,006,336 & 4.00 \\
Pinyin-token & 27,176 & 1,781,006,336 & 3.65 \\
Pinyin-source& 29,764 & 1,950,613,504 & 4.00 \\
\bottomrule
\end{tabular}
\caption{Pretraining budget per run. Passes are measured over the
training split after drop-last batching.}
\label{tab:pretraining-budgets}
\end{table}

\subsection{Downstream evaluation}
\label{app:downstream-evaluation}

\begin{table*}[!tp]
\centering
\small
\setlength{\tabcolsep}{4pt}
\begin{tabular}{@{}lcccrr@{}}
\toprule
Phenomenon & Chinese & Pinyin-token & Pinyin-source
& $\Delta_{\mathrm{tok}}$ & $\Delta_{\mathrm{src}}$ \\
\midrule
\multicolumn{6}{@{}l}{\emph{Functional licensing}} \\
FCI licensing
& $88.49 \pm 0.77$ & $48.82 \pm 6.09$ & $51.93 \pm 9.66$
& $+39.67$ & $+36.56$ \\
NPI licensing
& $71.61 \pm 6.96$ & $52.55 \pm 0.66$ & $52.09 \pm 2.62$
& $+19.05$ & $+19.52$ \\
\addlinespace
\multicolumn{6}{@{}l}{\emph{Constructional}} \\
Control/raising
& $84.36 \pm 2.87$ & $74.94 \pm 1.87$ & $69.97 \pm 3.20$
& $+9.42$ & $+14.39$ \\
Verb phrase
& $85.60 \pm 0.69$ & $73.49 \pm 1.44$ & $75.27 \pm 1.53$
& $+12.11$ & $+10.33$ \\
BA construction
& $74.90 \pm 1.00$ & $65.90 \pm 2.51$ & $68.80 \pm 3.53$
& $+9.00$ & $+6.10$ \\
\addlinespace
\multicolumn{6}{@{}l}{\emph{Anaphor}} \\
Anaphor
& $55.00 \pm 0.42$ & $33.24 \pm 1.36$ & $30.41 \pm 1.52$
& $+21.76$ & $+24.59$ \\
Anaphor*
& $57.33 \pm 1.79$ & $49.86 \pm 2.04$ & $45.61 \pm 2.29$
& $+7.47$ & $+11.72$ \\
\addlinespace
\multicolumn{6}{@{}l}{\emph{Argument structure}} \\
Argument structure
& $68.94 \pm 1.00$ & $63.89 \pm 0.81$ & $64.30 \pm 1.64$
& $+5.05$ & $+4.64$ \\
\addlinespace
\multicolumn{6}{@{}l}{\emph{Order-sensitive}} \\
Passive
& $79.24 \pm 2.37$ & $72.27 \pm 1.10$ & $72.36 \pm 0.94$
& $+6.97$ & $+6.88$ \\
Topicalization
& $80.47 \pm 2.20$ & $76.50 \pm 0.96$ & $74.64 \pm 2.46$
& $+3.97$ & $+5.83$ \\
Relativization
& $72.36 \pm 5.84$ & $68.70 \pm 3.44$ & $68.81 \pm 4.84$
& $+3.66$ & $+3.55$ \\
\addlinespace
\multicolumn{6}{@{}l}{\emph{Unstable categories}} \\
Nominal expression
& $50.37 \pm 4.76$ & $45.86 \pm 2.05$ & $44.47 \pm 3.93$
& $+4.52$ & $+5.90$ \\
Ellipsis
& $35.48 \pm 1.86$ & $36.41 \pm 1.38$ & $36.22 \pm 1.26$
& $-0.93$ & $-0.74$ \\
Question
& $58.72 \pm 1.48$ & $65.32 \pm 1.75$ & $67.90 \pm 4.33$
& $-6.59$ & $-9.18$ \\
Classifier
& $44.37 \pm 4.71$ & $44.89 \pm 1.84$ & $47.67 \pm 5.17$
& $-0.52$ & $-3.30$ \\
Quantifiers
& $36.72 \pm 15.59$ & $49.78 \pm 2.71$ & $44.83 \pm 3.75$
& $-13.05$ & $-8.11$ \\
\midrule
Overall
& $68.40 \pm 0.66$ & $61.19 \pm 0.83$ & $61.76 \pm 1.27$
& $+7.21$ & $+6.64$ \\
\bottomrule
\end{tabular}
\caption{ZhoBLiMP accuracy (\%), reported as mean $\pm$ sample
standard deviation across three training seeds. Anaphor includes all
anaphor pairs. Anaphor* excludes pairs that become identical under
Pinyin-Diacritic conversion. $\Delta_{\mathrm{tok}}$ and
$\Delta_{\mathrm{src}}$ give the Chinese mean minus the Pinyin-token
and Pinyin-source means, respectively, in percentage points.}
\label{tab:zhoblimp-full}
\end{table*}

\paragraph{Homophone-interference lexical probe.}
\leavevmode\phantomsection\label{app:homophone-probe}%
We construct 1,000 items from held-out Chinese Wikipedia sentences.
For each gold target word, we sample a distractor from the same
Pinyin-Toneless collision group
(Appendix~\ref{app:additional-overlap-results}). After tone marks
are restored, 229 pairs have identical Pinyin-Diacritic forms.
The \emph{Pinyin overlap*} subset contains the remaining 771 pairs.

Both controls reuse the same contexts and gold targets. The semantic
control uses a non-homophone distractor matched by semantic distance
to the target. The random control uses a randomly sampled
non-homophone. All control pairs remain distinguishable under
Pinyin-Diacritic conversion.

For each left context, the model scores two continuations: the gold target or the distractor, each followed by the original sentence suffix. An item is correct if the gold continuation receives the higher score.

\paragraph{ZhoBLiMP minimal pairs.}
\leavevmode\phantomsection\label{app:zhoblimp-full}%
We use the \texttt{train} split of the \href{https://huggingface.co/datasets/chinese-babylm-org/zhoblimp}{ZhoBLiMP dataset} with \texttt{all\_items=true} (35,400 minimal pairs).

For each minimal pair, we score both sentences in the model's input
representation. A model is correct if the acceptable sentence has a higher mean token-level log-probability than its counterpart.
Table~\ref{tab:zhoblimp-full} gives the
scores for all 15 phenomena, grouped by the classes used in
Table~\ref{tab:zhoblimp-typology}. Each class score is the
macro-average over its constituent phenomena. For the anaphor class,
the main table uses the filtered Anaphor* subset.

\paragraph{Single-character final matching.}
\leavevmode\phantomsection\label{app:phonological-probe}%
Each item contains a query character, a candidate with the same
Pinyin final, and a distractor with a different final. We extract
finals with \texttt{pypinyin.Style.FINALS} and \texttt{strict=True},
ignoring tone. We exclude numerals, characters with fewer than
500 occurrences, and characters with ambiguous or missing readings.
The two candidates are approximately matched in frequency.

Of 1,000 items constructed from held-out Chinese Wikipedia text,
670 have two candidate answers that each occupy one token under
both tokenizers. We sample 600 of these items with seed~42,
covering 397 distinct query characters and 34 finals. The gold
answer occupies the first position in 296 items and the second in
304. All nine frozen checkpoints are evaluated on the same items.

For the query character \emph{zhū}, the Chinese prompt reads:
\begin{quote}
\begin{CJK}{UTF8}{gbsn}
不看声调。\allowbreak{}与诸韵母相同的是哪个字？\allowbreak{}答案：
\end{CJK}\\
\emph{Ignore tone. Which character has the same final as
zhū? Answer:}
\end{quote}
For Pinyin models, we convert the entire prompt to tone-marked
Pinyin. We score each candidate separately as a continuation of
the prompt, without displaying the candidates or A/B labels.
Given prompt \(x\), we select the answer token \(a\) with the higher
\(\log p_\theta(a\mid x)\). Each Pinyin answer continuation includes
a leading space.

Zero-shot prompts contain only the test query and instruction.
Four-shot prompts prepend four answered examples drawn from the
original 1,000-item pool. For each test item, we exclude examples
sharing any character with that item. Five distinct demonstrations
are used across all prompts. The demonstration pool overlaps the
evaluation set, but no item serves as its own demonstration.
Table~\ref{tab:phonological-probe} reports accuracy for both
prompting conditions.

\section{Software and Implementation Details}
\label{app:implementation-details}

\paragraph{Tokenizers.}
We train superBPE \citep{liu-etal-2025-superbpe} using
the \href{https://github.com/PythonNut/superbpe}{official implementation}
and its
\href{https://github.com/alisawuffles/tokenizers-superbpe}
{tokenizers-superbpe dependency} (upstream commit \texttt{757f2a5}).
In our local copy, we change the trainer to skip tokens spanning
more than 32 words, up from the upstream limit of 4.
For the comparison in Table~\ref{tab:fertility}, we load the
\href{https://huggingface.co/Qwen/Qwen-7B}{Qwen-7B tokenizer}
with \texttt{AutoTokenizer.from\_pretrained} and
\texttt{trust\_remote\_code=True}. We count tokens with
\texttt{add\_special\_tokens=False}.

\paragraph{Semantic-distance models.}
For the dispersion analysis in
Appendix~\ref{app:additional-semantic-dispersion}, we use
sentence-transformers with
\href{https://huggingface.co/BAAI/bge-small-zh-v1.5}
{\texttt{BAAI/bge-small-zh-v1.5}} and
\href{https://huggingface.co/shibing624/text2vec-base-chinese}
{\texttt{shibing624/text2vec-base-chinese}}.
We L2-normalize the embeddings and compute cosine distance as
\(1-\cos(\mathbf{x},\mathbf{y})\). Random baseline sampling uses
seed 20260430. For the semantic control in
Appendix~\ref{app:homophone-probe}, we use the text2vec model
with normalized embeddings and a batch size of 128.

\paragraph{Software versions.}
Unless otherwise stated, we use pypinyin 0.55.0, tokenizers 0.22.2,
transformers 4.57.6, datasets 4.5.0, sentence-transformers 5.1.2,
PyTorch 2.8.0, NumPy 2.0.2, scikit-learn 1.6.1, and tqdm 4.67.3.

\end{document}